\documentclass[10pt,twocolumn,letterpaper]{article}

\usepackage{iccv}
\usepackage{graphicx}
\usepackage{times}
\usepackage{epsfig}
\usepackage{graphicx}
\usepackage{amsmath}
\usepackage{amssymb}

\usepackage[pagebackref=true,breaklinks=true,letterpaper=true,colorlinks,bookmarks=false]{hyperref}

\iccvfinalcopy

\newcommand*{\ShowNotes}{}

\usepackage[usenames]{color}
\definecolor{darkred}{rgb}{0.7,0.1,0.1}
\definecolor{darkgreen}{rgb}{0.1,0.7,0.1}
\definecolor{cyan}{rgb}{0.0,0.7,0.7}
\definecolor{dblue}{rgb}{0.2,0.2,0.8}
\definecolor{maroon}{rgb}{0.76,.13,.28}
\definecolor{burntorange}{rgb}{0.81,.33,0}
\definecolor{purple}{rgb}{0.7,0,0.7}

\ifdefined\ShowNotes
  \newcommand{\colornote}[3]{{\color{#1}\bf{#2: #3}\normalfont}}
\else
  \newcommand{\colornote}[3]{}
\fi

\def\iccvPaperID{4597}

\ificcvfinal\fi

\begin{document}

\title{{\em k-NNN:} Nearest Neighbors of Neighbors for Anomaly Detection}
\author{Ori Nizan\\
Technion Israel institute of technology\\
Institution1 address\\
{\tt\small snizori@campus.technion.ac.il}
\and
Ayellet Tal\\
Technion Israel institute of technology\\
First line of institution2 address\\
{\tt\small ayellet@ee.technion.ac.il}
}

\maketitle

\ificcvfinal\thispagestyle{empty}\fi

\maketitle

\begin{abstract}
Anomaly detection aims at identifying images  that deviate significantly from the norm.
We focus on algorithms that embed the normal training examples in space and when given a test image, detect anomalies based on the features' distance to the k-nearest training neighbors.
We propose a new operator that takes into account the varying structure \& importance of the features in the embedding space.
Interestingly, this is done by taking into account not only the nearest neighbors, but also the neighbors of these neighbors (k-NNN).
We show that by simply replacing the nearest neighbor component in existing algorithms by our k-NNN operator, while leaving the rest of the algorithms untouched, each algorithms' own results are improved.
This is the case both for common homogeneous datasets, such as flowers or nuts of a specific type, as well as for more diverse datasets.
\end{abstract}

\section{Introduction}
\label{sec:Introduction}

Anomaly detection aims at finding patterns in the data that do not conform to the expected "behavior"~\cite{an2015variational}.
It has numerous applications, most notably in  manufacturing, surveillance, fraud detection, medical diagnostics, autonomous cars, and 
detecting outliers.
Out of the variety of anomaly detection 
methods~\cite{caron2021emerging,chen2020simple,chen2021exploring,targ2016resnet},
we focus on those that rely on the {\em k-Nearest-Neighbor (k-NN)} operator~\cite{bergman2020deep,Cohen2020SubImageAD,Gu2019StatisticalAO,reiss2021panda,reiss2021mean}. 
These methods learn the embedding of {\em normal} images or patches.
Given a test image, the distance of its embedding to its k-nearest (training) neighbors is computed and this distance determines whether the test image is anomalous or not.
The underlying assumption is that anomalous features  should reside farther away from normal features than  normal features from each other. 
Thus, a point is considered anomalous when its average distance to its $k$ nearest neighbors  exceeds a certain threshold.

\begin{figure}[t]
    \centering
  \centering
    \begin{tabular}{cc}
    \includegraphics[width=0.23\textwidth]{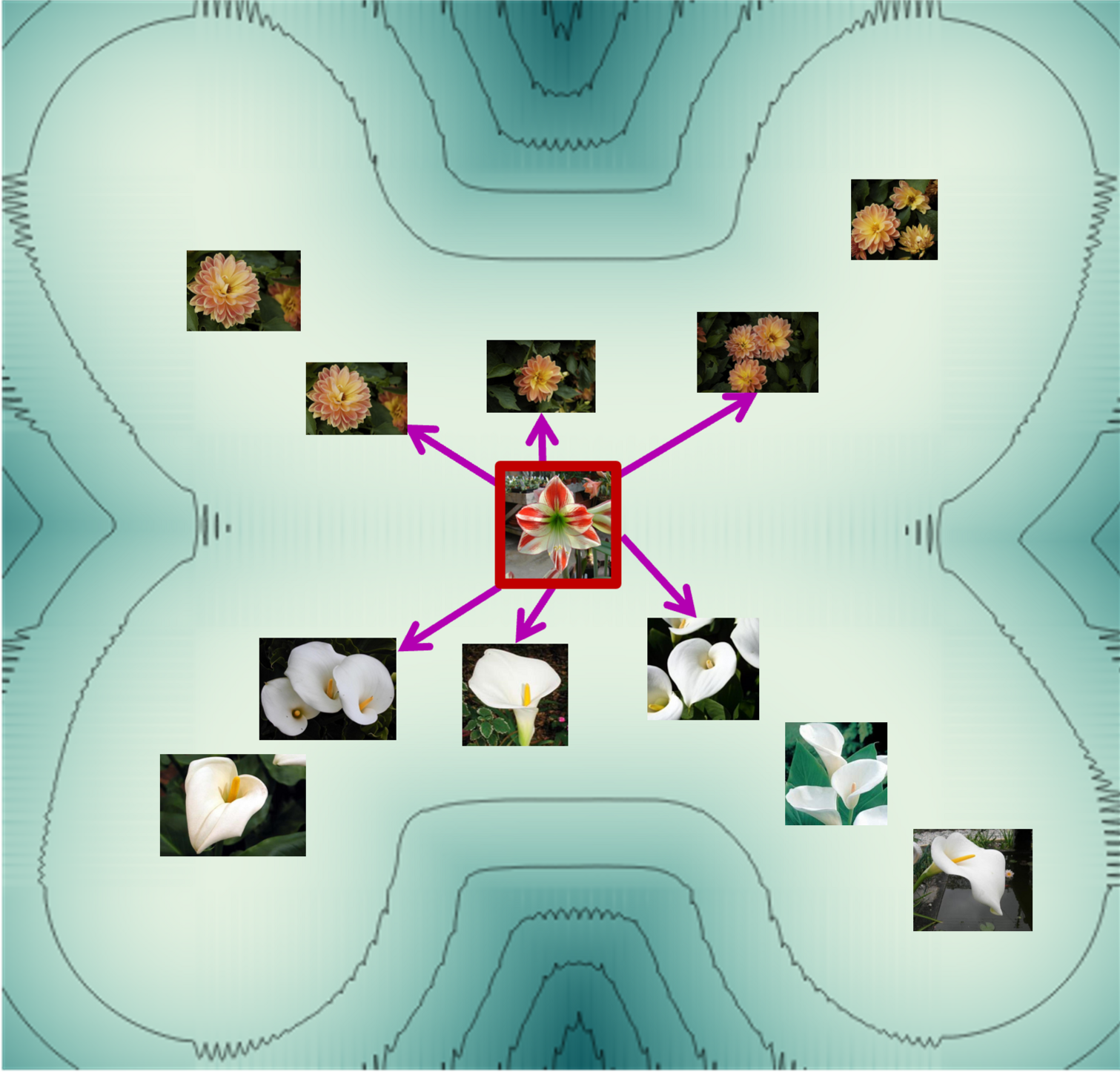} & 
    \includegraphics[width=0.23\textwidth]{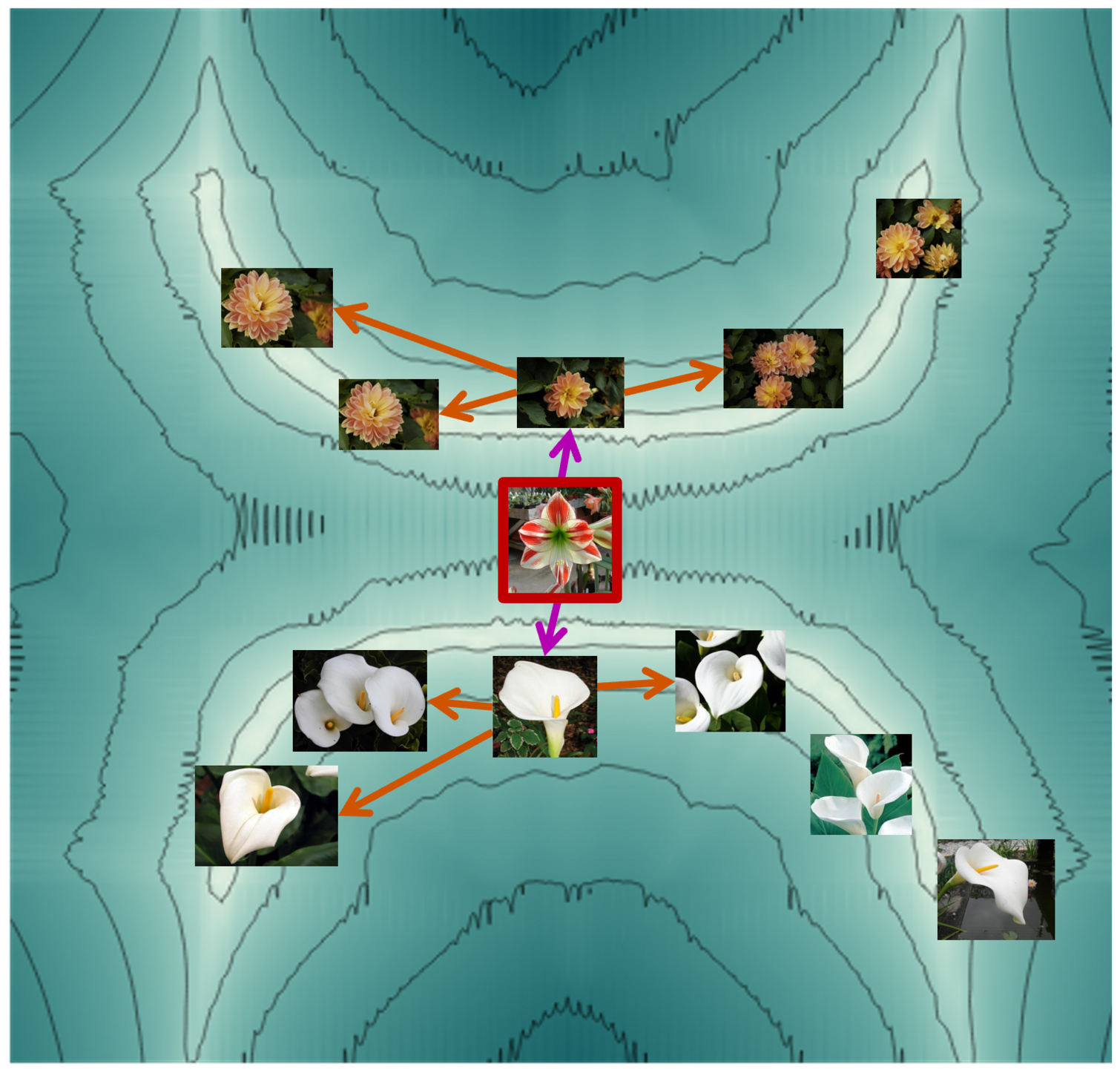} \\
   (a) {\em  k-NN} & (b) {\em k-NNN} \\
    \end{tabular}
  
    \caption{
    {\bf Neighbors of neighbors illustration.}
 A common approach is to base the anomaly score on the distances between a test image  to its k-nearest neighbors ({\em k-NN}) in the feature space.
 In this figure, the yellow and the white flowers are considered normal, whereas the flower in the red rectangle is anomalous.
 The background color represents the anomaly score.
 (a) Since the normal set is diverse,  {\em k-NN}-based methods might fail, since the {\color{purple}{distances}}  of the anomalous image to its neighbors is smaller than the distances between similar images to each other (e.g., between the white flowers). 
 (b) Our {\em k-NNN} operator sets better {\color{burntorange}{distances}} between neighbors, which reflect the  diversity of the normal examples.
 It will correctly detect the anomalous flower as such.
  }
  \label{fig:teaser}
\end{figure}

A major disadvantage of this approach is that the structure and the importance of the features in the embedding space is not taken into account when looking for anomalies.
Figure~\ref{fig:teaser} shows such a case, in which the normal set varies and consists of flowers that belong to different classes, with different inner distances (the flowers are not be classified beforehand).
A flower that does not resemble any of the normal flowers (in the red rectangle) will not be detected as anomalous by the {\em k-NN} operator, because its distance to its nearest flowers is less than the distances between normal flowers in different regions of the embedding space.
Our operator, which implicitly takes the structure of the  space  into account, will detect this flower as anomalous.

\begin{figure}[t]
\centering
\begin{tabular}{cc}
\includegraphics[width=0.22\textwidth]{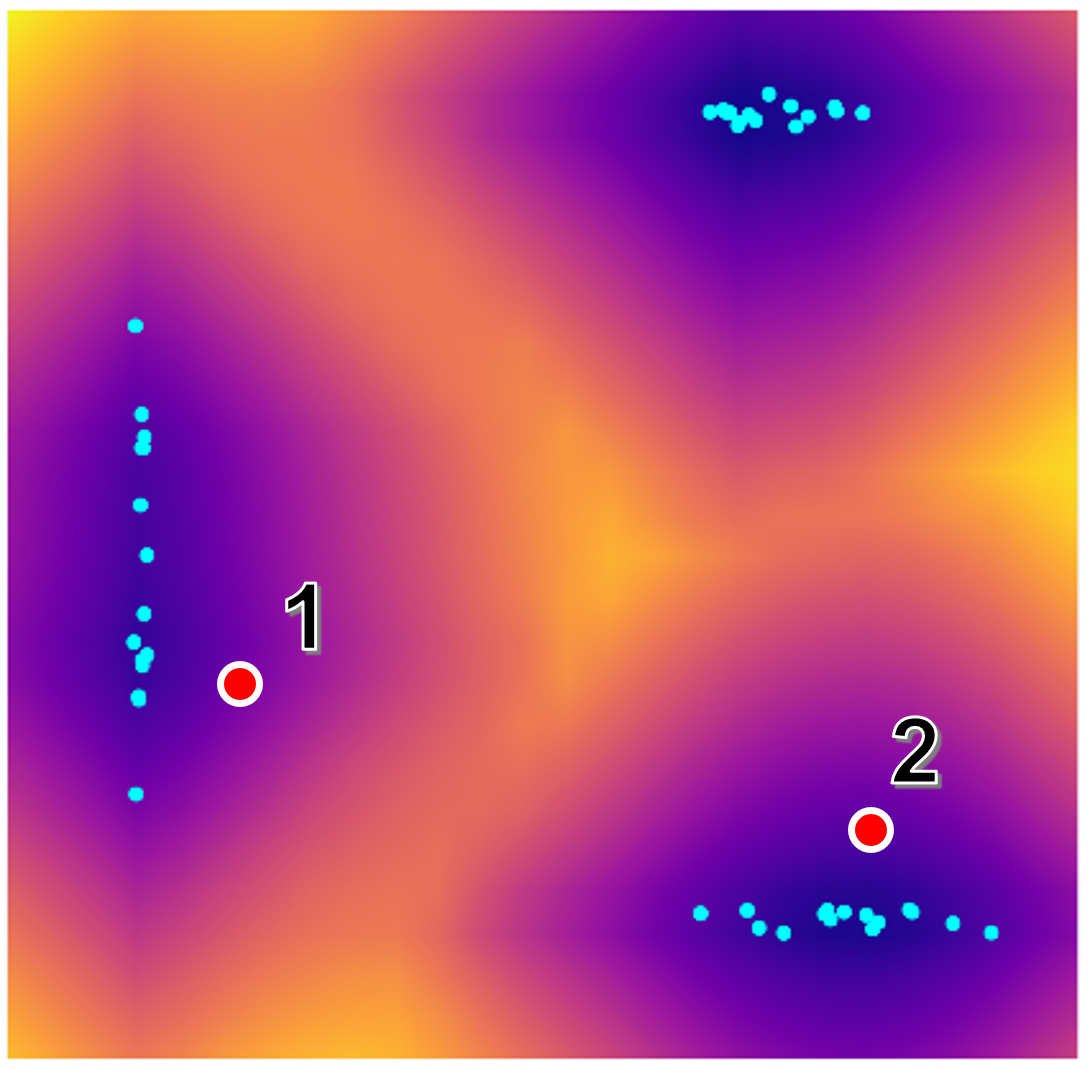} &
\includegraphics[width=0.22\textwidth]{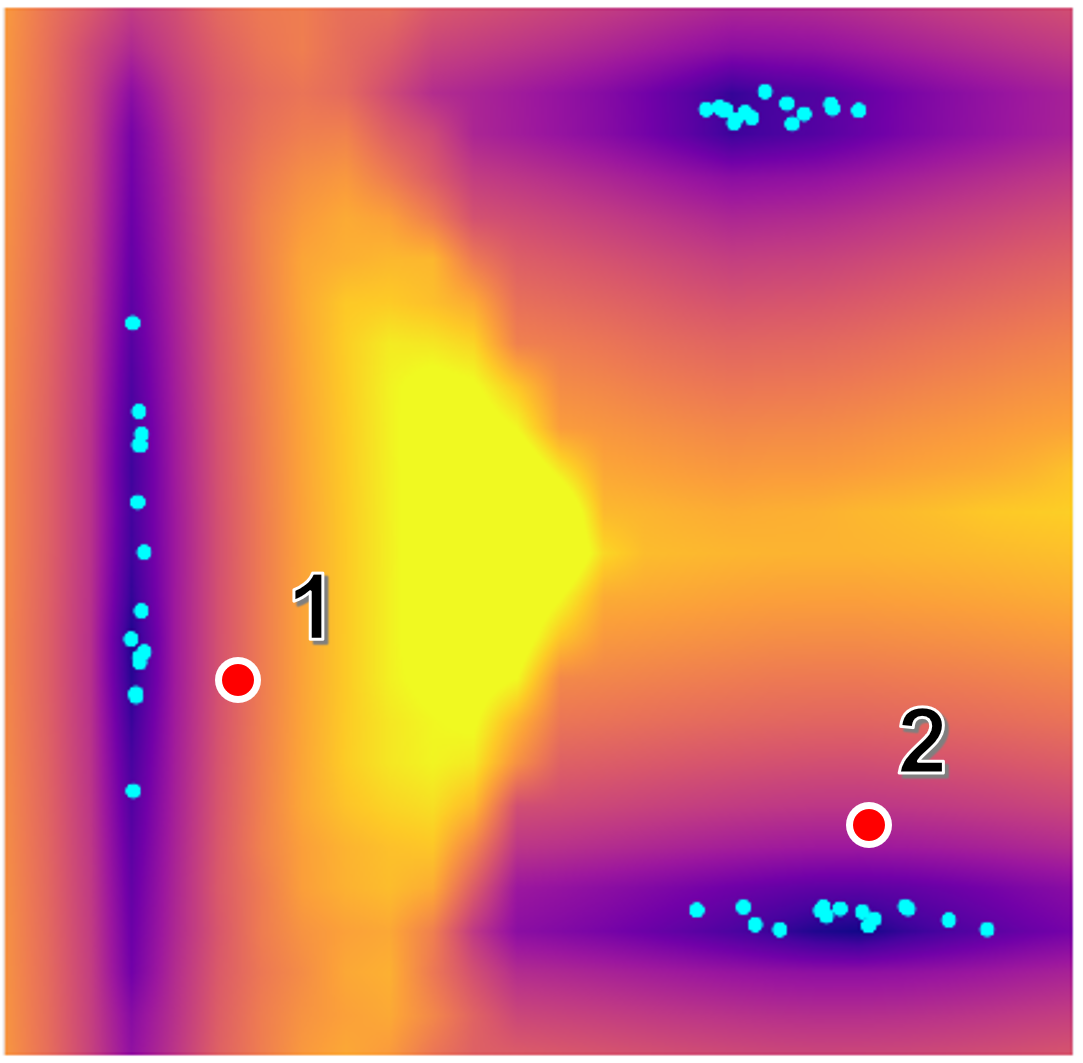} \\
(a) {\em k-NN}: $1$,$2$ are normal & (b)  {\em k-NNN}: $1$,$2$ are anomalous \\
\end{tabular}
\caption{
{\bf {\em k-NNN} benefit.} 
The {\color{cyan}{cyan points}}  represent the 2D embedding of normal images.
The heat maps show the {\em 5-NN} distance of each point on the plane; the yellower a region, the more anomalous it is. 
The distance between the {\color{red}{red anomalous points}}
to their $5$-nearest {\color{cyan}{cyan neighbors}} is equal to the distances between the cyan points themselves.
Thus,  the classical {\em k-NN} operator  fails to detect them as anomalous. 
Differently, our {\em k-NNN} operator, which uses the neighbors statistics, detects them correctly.  
}
\label{fig:illustration}
\end{figure}

The structure of the embedding space is important also when the normal set is homogeneous, as illustrated in Figure~\ref{fig:illustration} for a synthetic example. 
The 2D embedding of the {\color{cyan}{normal training points}} 
lie on three lines, two of which are parallel and one is perpendicular to them.
Two {\color{red}{anomalous points}}, marked as $1$ and $2$ (in red), lie above the horizontal line and to the right of the vertical line, respectively.
Their {\em 5-NN} distance is the same as that of the normal points between themselves, and thus they might not be identified as anomalous by {\em k-NN}-based methods. 
Similar cases are likely to occur when there are not enough training samples.

We propose a novel operator, termed the {\em k-nearest neighbors-of-neighbors (k-NNN)}, which addresses this problem, as illustrated in Figures~\ref{fig:teaser}-\ref{fig:illustration}.
It  differentiates between regions and considers the more indicative features at certain regions as more influential.
For instance, in Figure~\ref{fig:illustration} the feature that makes point~$2$ anomalous is its $y$~feature, whereas the feature that makes point~$1$ anomalous is its $x$~feature.
We show how to efficiently realize this idea of considering regions differently,  by simply looking at the neighbors of neighbors of a test point.
Intuitively, the neighbors of neighbors provide information about regions, which balances between a global view of the dataset and a more local view, which is based only on the immediate neighbors.

To consider the feature importance, we need to find the directions associated with the anomalies.
The classical {\em Principal component analysis (PCA)}  analyzes datasets of high-dimension features, while preserving the maximum amount of information. 
We observe that for anomaly detection,  Eigen vectors associated with small Eigen values matter more than those of large values.
Furthermore, as shown in Figure~\ref{fig:illustration}, the difference between an anomalous point and its nearest neighbor(s) is perpendicular to the direction of the large Eigen vector(s). 
Intuitively, this is so since anomalies are characterized by features not present in the dataset.
We show how to utilize this observation within our operator.

To demonstrate the benefit of our approach, we replace the {\em k-NN} operator used in several anomaly detection algorithms
with our {\em k-NNN}  operator.
We show how this modification manages to improve the results of each algorithm on a variety of datasets.

Hence, this paper makes two contributions:
\begin{enumerate}
    \item 
    It introduces a novel, general, efficient and accurate operator---the {\em k-NNN} operator,
    which provides an "in-between" look at the data, between local and global.
It benefits both diverse and homogeneous normal  sets.
\item
It proposes a novel normalization scheme, which gives more weight to the small Eigen values and copes with the challenge of having small datasets.
\end{enumerate}

\section{Related work}
\label{sec:related}

\noindent
{\bf Anomaly detection.}
Anomaly detection is important to discover potentially dangerous situations, in the manufacturing industry for detecting product faults, in medicine for diagnosing diseases etc. 
It is a highly challenging task  due to image structure, varying environmental conditions, 
imbalanced datasets, and data diversity. 
 Hence, this task has attracted a huge amount of research.
We refer the reader to a couple of comprehensive and excellent surveys~\cite{chalapathy2019deep,tran2022anomaly}. 

Hereafter, we consider methods that detect whether or not  an image is anomalous and do not aim to segment it.
They may be categorized to three classes, as follows.

{\em Reconstruction-based} methods learn a set of basis functions on the training data.
Given a test image, they attempt to reconstruct it using these functions. 
If the test image cannot be reconstructed, it is considered anomalous.
The set of basis functions vary.
Examples include 
K-means~\cite{a_k_means_clustering_algorithm},
K nearest neighbors (k-NN) \cite{eskin2002geometric},
principal component analysis (PCA)~\cite{abdi2010principal} etc.
Deep learning has been used as well~\cite{sakurada2014anomaly, zhang2016colorful}.

{\em Distribution-based} methods  model the probability density function (PDF) of the distribution of the normal data~\cite{ eskin2002geometric,lin2017feature}.
Given a test example, it is  evaluated using the PDF.
If the probability is small, it is considered anomalous. 
Deep learning can be applied as well~\cite{li2018anomaly,zong2018deep}.

{\em Classification-based methods} are the most prevalent recently. 
They includes one-class methods~\cite{ruff2018deep,scholkopf1999support, tax2004support} and
self-supervised learning~ \cite{Bergman2020Classification_Based,gidaris2018unsupervised,golan2018deep,  hendrycks2018deep}.
Recently it was shown in  ~\cite{bergman2020deep} that a simple method, which is based on {\em k-NN},  outperforms such self-supervised methods.

Nearest neighbors, which we pursue in this paper, may be considered as reconstruction-based or as distribution-based, since it performs density estimation.

\begin{figure*}[t]
\centering
\begin{tabular}{cccc}
\includegraphics[width=0.24\textwidth]{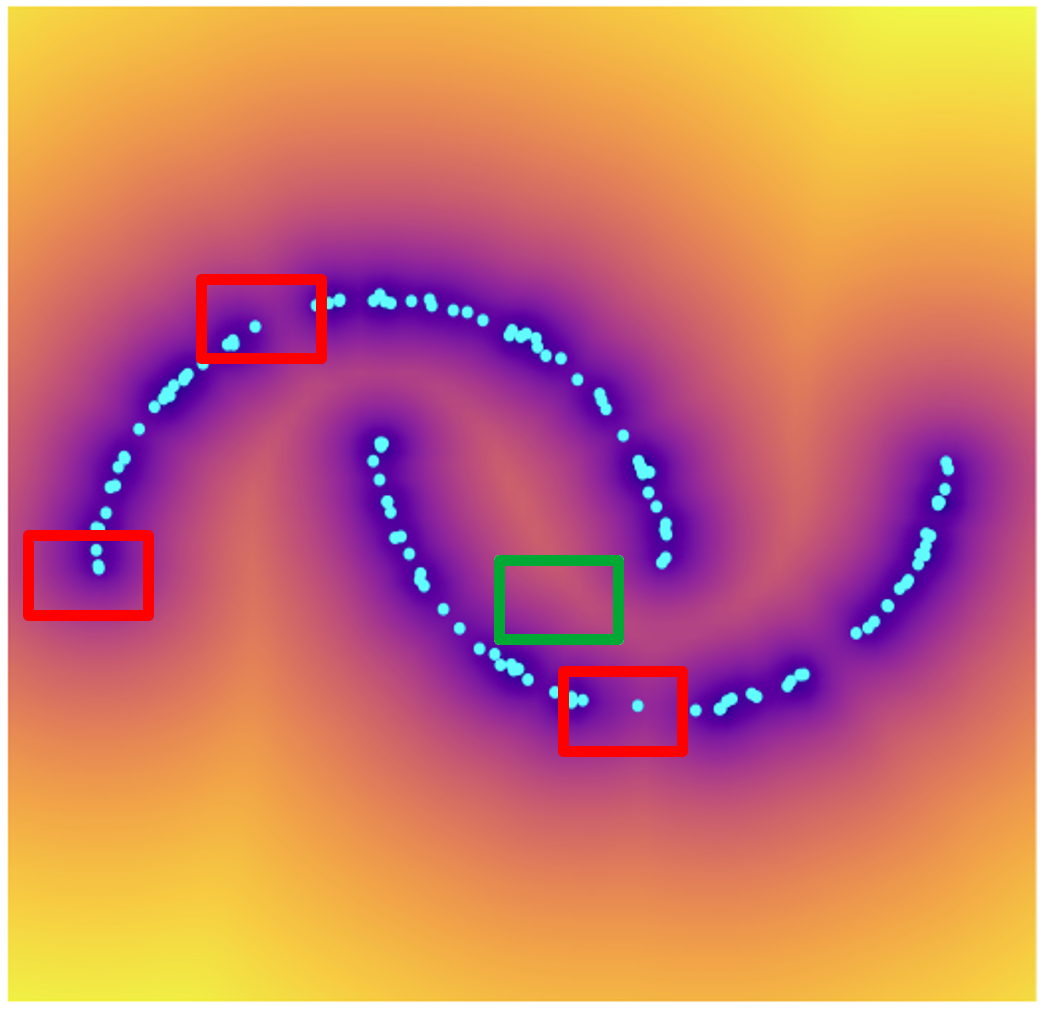} &
\includegraphics[width=0.24\textwidth]{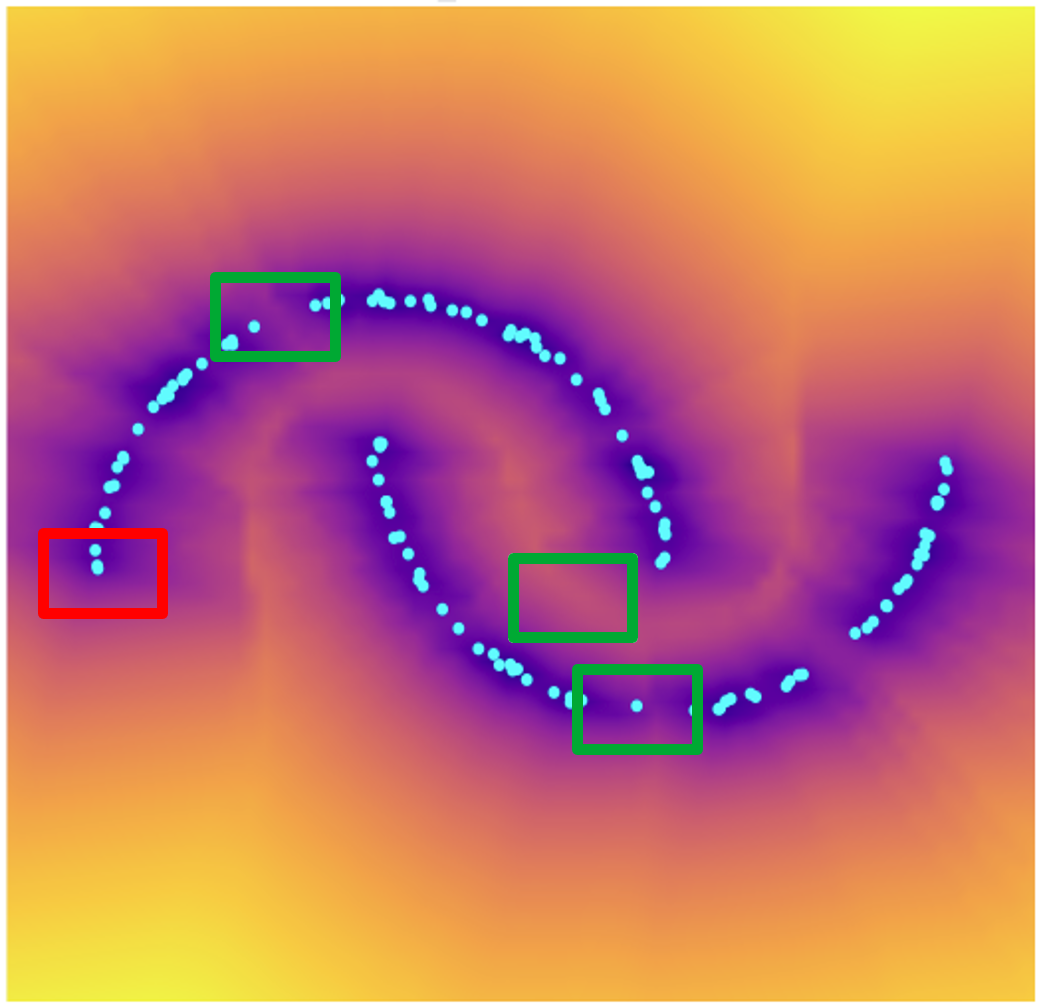} &
\includegraphics[width=0.24\textwidth]{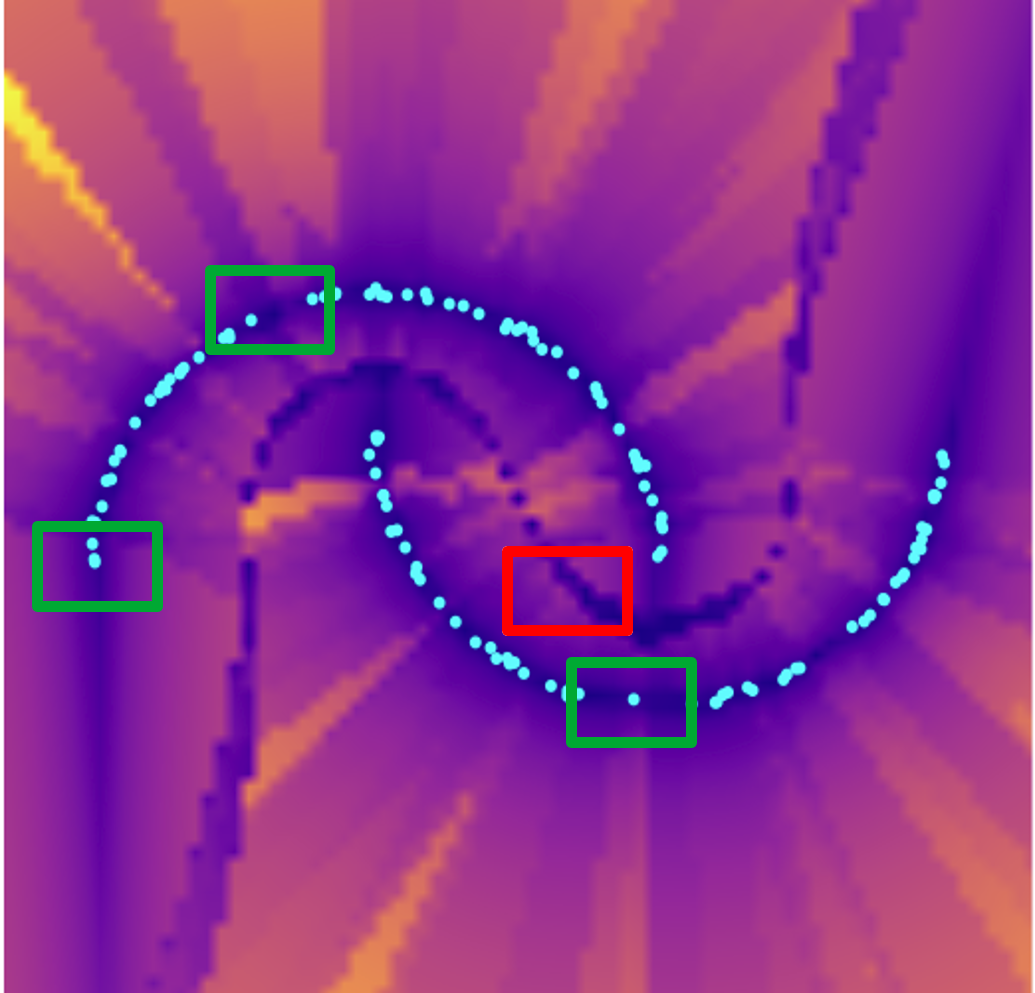} &
\includegraphics[width=0.24\textwidth]{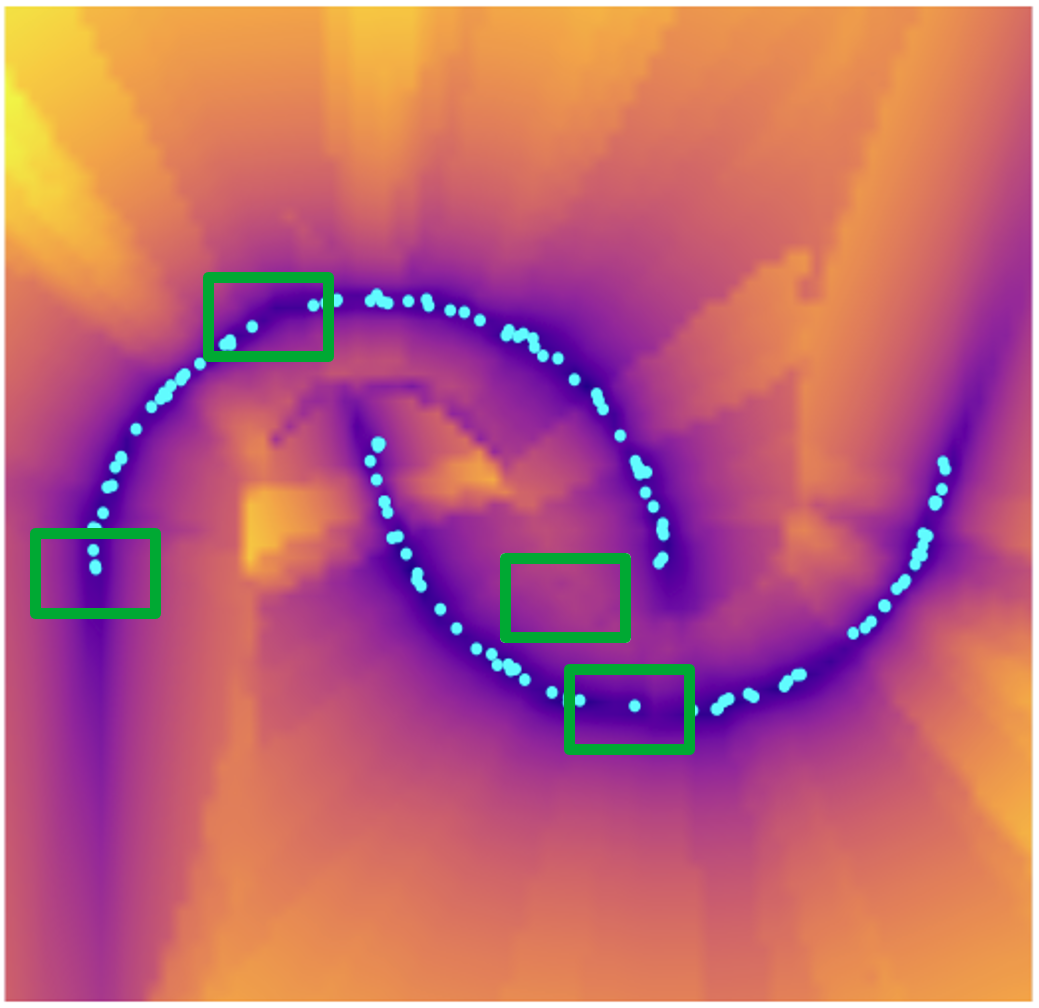} \\ 
(a) {\em k-NN} & (b) {\em Global} &
 (c) {\em Local} & (d) {\em k-NNN}\\
\end{tabular}
\caption{
{\bf Different types of normalization.}
The {\color{cyan}{cyan points}} represent the {\color{cyan}{normal points}} in a 2D embedding space; they lie along two circular arcs.
We expect that all the normal points will reside along these arcs, even if  the arcs contain holes or terminate; we also expect that points that lie in other regions of the plane will be anomalous.
The background color represents the anomaly score of each region according to the specific method: The more yellowish a region, the more anomalous it is.
While our {\em k-NNN}~(d) correctly classifies the plane (blue regions are only along the arcs), the local method erroneously considers as normal the region in-between the arcs~(c), and the global and {\em k-NN} method erroneously detect regions along the arcs (in holes or beyond termination) as anomalous~(a-b).
In this figure  {\color{darkgreen}{green rectangles}} mark  {\color{darkgreen}{correct outcome}} (normal/anomalous) and  {\color{red}{red rectangles}} mark {\color{red}{incorrect outcome}}.}
\label{fig:dots_normalization}
\end{figure*}

\vspace{0.05in}
\noindent
{\bf The k-NN operator.}
Nearest neighbor search has been utilized across a wide range of applications in computer vision.
The {\em k-NN} operator 
has been found beneficial in classification and correspondence~\cite{Arbel2019Partial,boiman2008defense,tan2006effective}, 
intrusion detection~\cite{liao2002use},
medical applications~\cite{li2012using,medjahed2013breast},
fault detection~\cite{bergman2020deep,reiss2021panda} and more.
In some applications approximation of the {\em k-NN} operator was studied \cite{jegou2010product,malkov2018efficient,rajaraman2011mining}.
We focus, however, on the exact {\em k-NN} operator in the context of anomaly detection.
The most related works to ours are~\cite{Cohen2020SubImageAD,gong2019memorizing,norlander2019latent,pang2020deep,reiss2021panda,roth2021towards,winkens2020contrastive}, which use nearest neighbors for anomaly detection.

\section{Method}
\label{sec:model}

Given an image, our goal is to determine whether it is anomalous or not. 
This should be done in a semi-supervised manner, utilizing only a dataset of normal images, without anomalies.
We follow the approach in which features are extracted during training, in order to represent normal images.
During inference, a given test image is passed through the feature extractor and the  $k$-nearest neighbors in the (training) feature space are found.
An anomaly score is derived from the distances to these  nearest neighbors. 

\vspace{0.05in}
\noindent
{\bf Eigen-vector for anomaly detection.}
In order to take into account the shape of the embedding space, we estimate the space directions using its Eigen vectors.
Recall that the greater the Eigen value, the larger the variance of the data in the direction of the corresponding Eigen vector.
Our proposed normalization is based on our observation that small Eigenvalues tend to correspond to anomaly directions more than large Eigenvalues.
This can be explained by the fact that a small variation means that normal images are close in that direction~\cite{rippel2021modeling}.
Thus, a small deviation in this direction is more likely to be an anomaly.

\begin{figure*}[tb]
\centering
\includegraphics[width=0.95\textwidth]{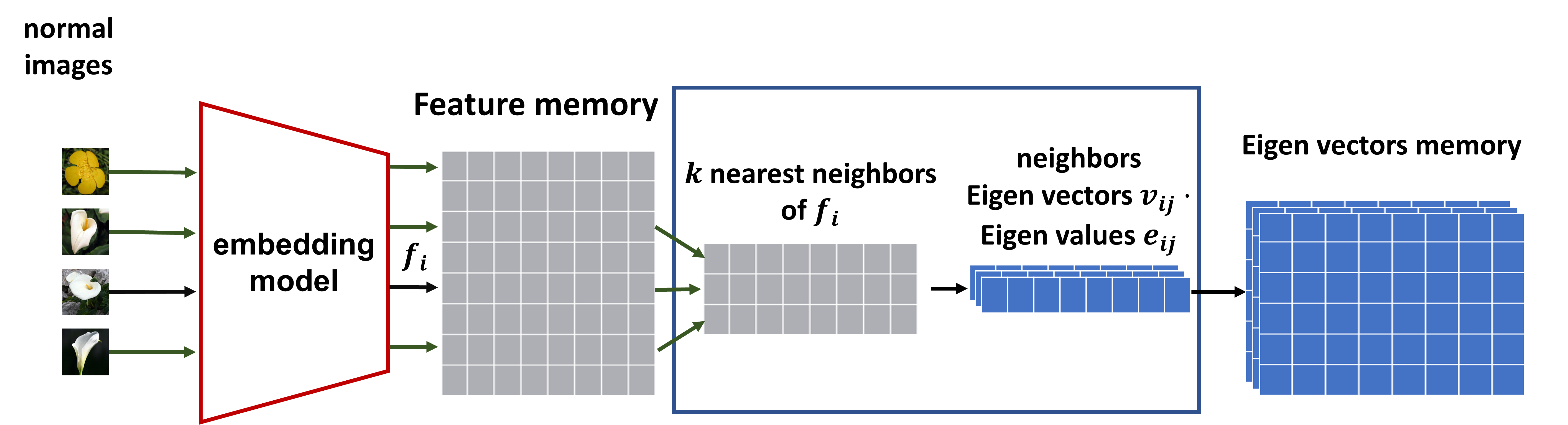}
\caption{
{\bf k-NNN training.}
Given training normal images, their embeddings, $f_i$, are computed.
Then, the nearest neighbors of each image embedding is computed. 
The Eigen vectors and Eigen values, derived from their $k$ neighbors, are computed and stored.
}
\label{fig:preprocessing}
\end{figure*}

\begin{figure*}[tb]
\centering
\includegraphics[width=0.95\textwidth]{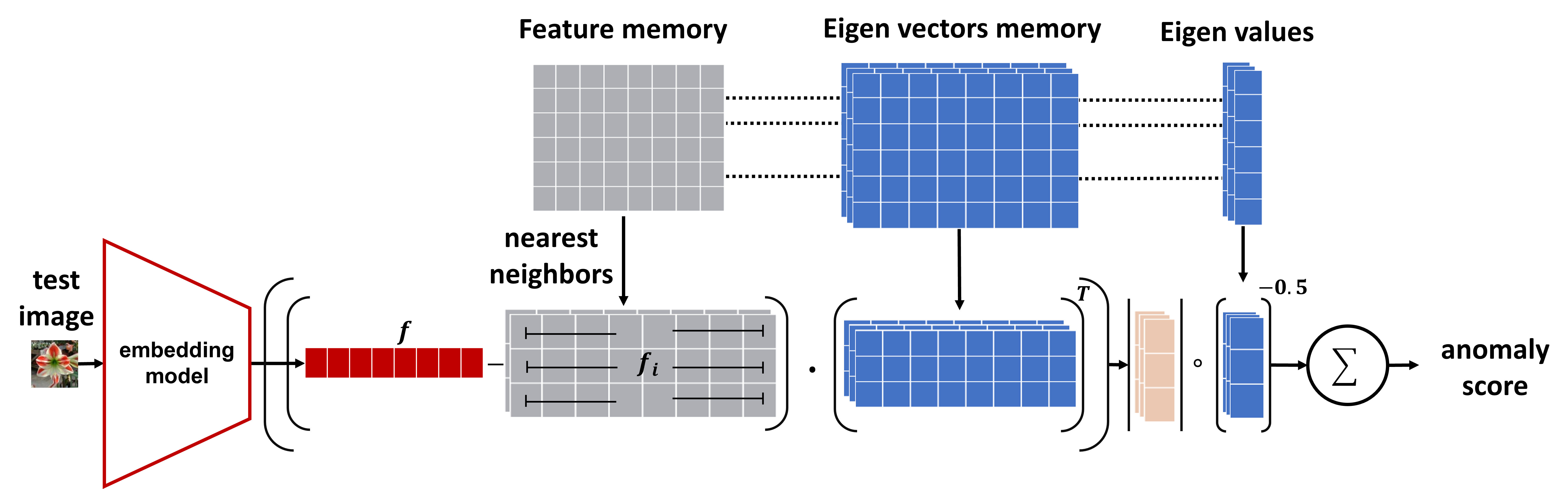}
\caption{
{\bf {\em k-NNN} inference.} 
Given an input image, its embedding $f$ is first computed.
Its $k$ nearest neighbors are found
and the Eigen values \& vectors of each neighbor are extracted from the memory.
An  anomaly score is calculated according to Eq.~\ref{eq:testf}.
}
\label{fig:inference}
\end{figure*}

\vspace{0.05in}
\noindent
{\bf The neighbors of neighbors operator.}
One may consider a couple of setups.
In a global setup, the Eigen vectors are determined for the whole training (normal) set  during pre-processing.
In a local setup, the Eigen vectors are calculated for a  test point based on its $k$-nearest neighbors in the training set.
We propose an "in-between" operator, which
 gathers more statistical information on the embedding space than the local operator and not as much as the global operator.
In particular, for each neighbor we utilize the Eigen vectors (/values) of its neighbors.
We elaborate on the realization of this idea hereafter.

Figure~\ref{fig:dots_normalization} illustrates the intuition behind our operator.
The  {\color{cyan}{normal points}}, in cyan,  lie along one of two circular arcs.
Obviously, normal points should lie along these arcs, even in holes and beyond the arcs' termination, whereas anomalous points should reside elsewhere in the plane.
In this figure, the plane is colored according to its normality/anomaly, as determined by each method.
Blue regions are considered to be normal by the method, whereas yellow regions are considered anomalous.
The {\color{darkgreen}{green rectangles}} highlight regions where the specific method {\color{darkgreen}{correctly classifies}} points as normal or anomalous. 
The {\color{red}{red rectangles}} highlight regions
in which the specific method {\color{red}{fails to  classify}} points. 
It can be seen that our method enjoys the benefits of all worlds---global \& local.
This result is analyzed and supported quantitatively  in Section~\ref{subsec:ablation}.

To realize our operator, during training (Figure~\ref{fig:preprocessing}), we first compute the feature vector of each  training image, using any preferable embedding model.
Then,  we compute the $k$ nearest neighbors in feature space for each point of the training data.
From these neighbors, we compute the point's $n$ Eigen vectors and their corresponding Eigen values and store this information. 
Hence, the Eigen vectors (/values) are relative to each individual training point, regardless of the test point.

At inference (Figure~\ref{fig:inference}), given a test point and its feature vector, $f$, we find its $k$ nearest neighbors among the training samples, $f_i$, $1\leq i \leq k$. 
Each of these $f_i$s is already associated with $n$ Eigen vectors and Eigen values, $v_{ij}$ and $e_{ij}$, $1\leq j\leq n$, computed during training.

Following our observation, for a point to be considered normal,  the difference vector between it and its neighbor should be parallel to the large Eigen vectors
(parallel to the distribution of the normal embeddings).
Reversely, for a point to be considered anomalous, this vector  should be perpendicular to the large Eigen vectors.
Thus, we calculate the anomaly score {\em AS} of a feature vector $f$ as follows.
\begin{equation}
    AS(f) = \sum_{i=1}^{k}\sum_{j=1}^{n}{|(f-f_{i})\cdot v_{ij}| \cdot \frac{1}{\sqrt{{e_{ij}}} }}.
    \label{eq:testf}
\end{equation}
In Eq.~\ref{eq:testf}, 
the difference between the test feature vector and that of its neighbor  is multiplied by the different Eigen vectors,  specific to the $i^{th}$ nearest neighbor.
The more parallel these vectors are, the larger the value of this multiplication.
Furthermore, this number is multiplied by square root of the inverse of the Eigen value, giving more weight to the small Eigen values.
Figure~\ref{fig:dots_normalization}(d) demonstrates that the {\em k-NNN} operator indeed classifies the plane properly.

\begin{figure*}[t]
\centering
\begin{tabular}{cc}
\includegraphics[width=0.25\textwidth]{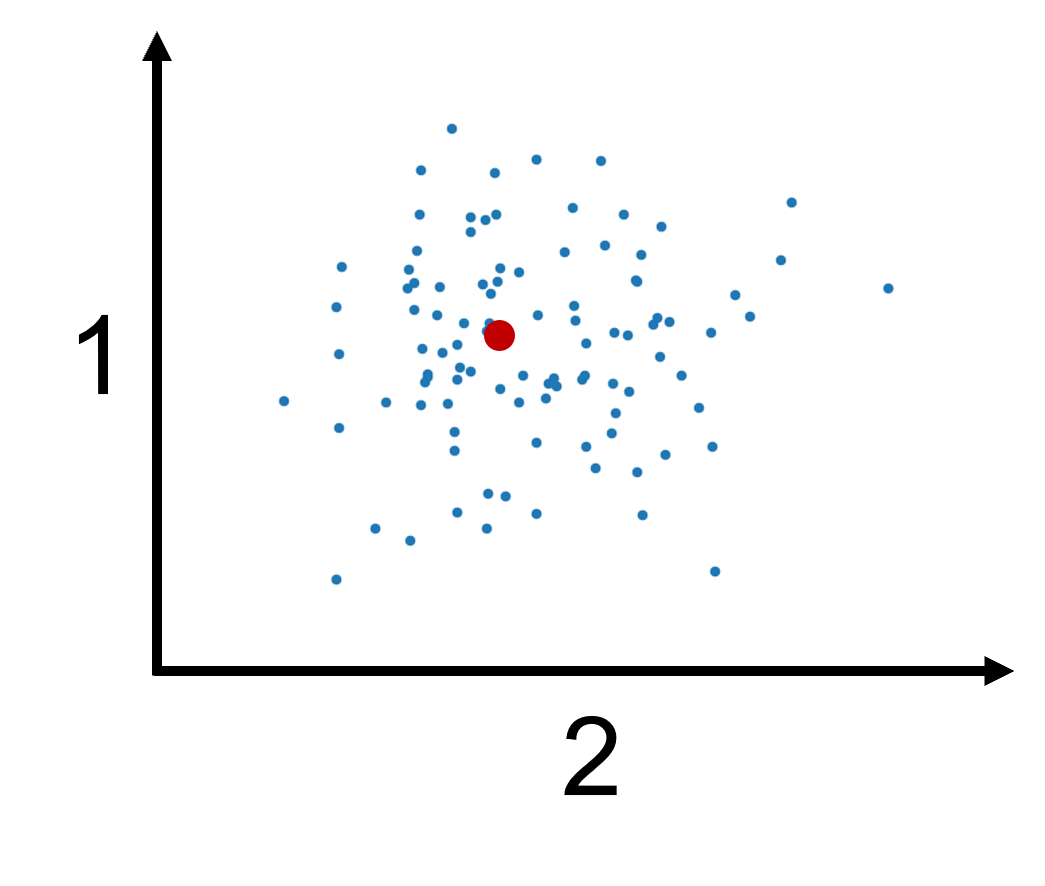} 
\includegraphics[width=0.25\textwidth]{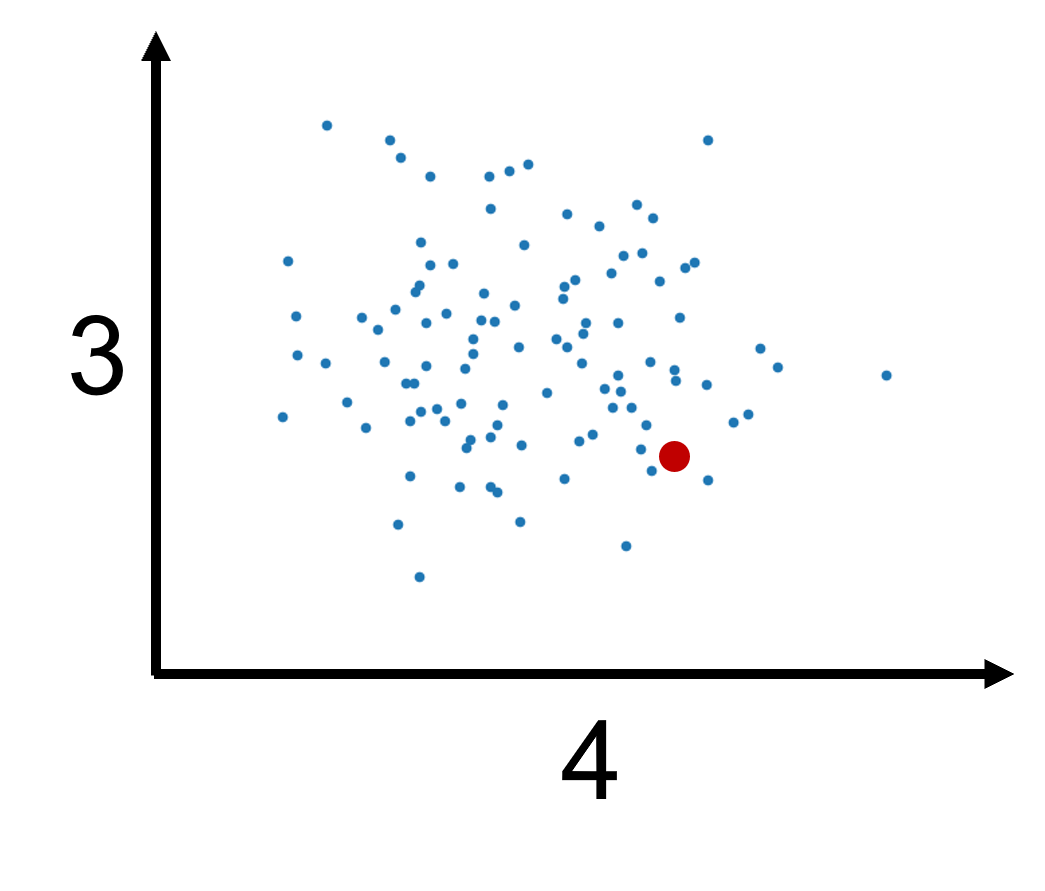} &
\includegraphics[width=0.25\textwidth]{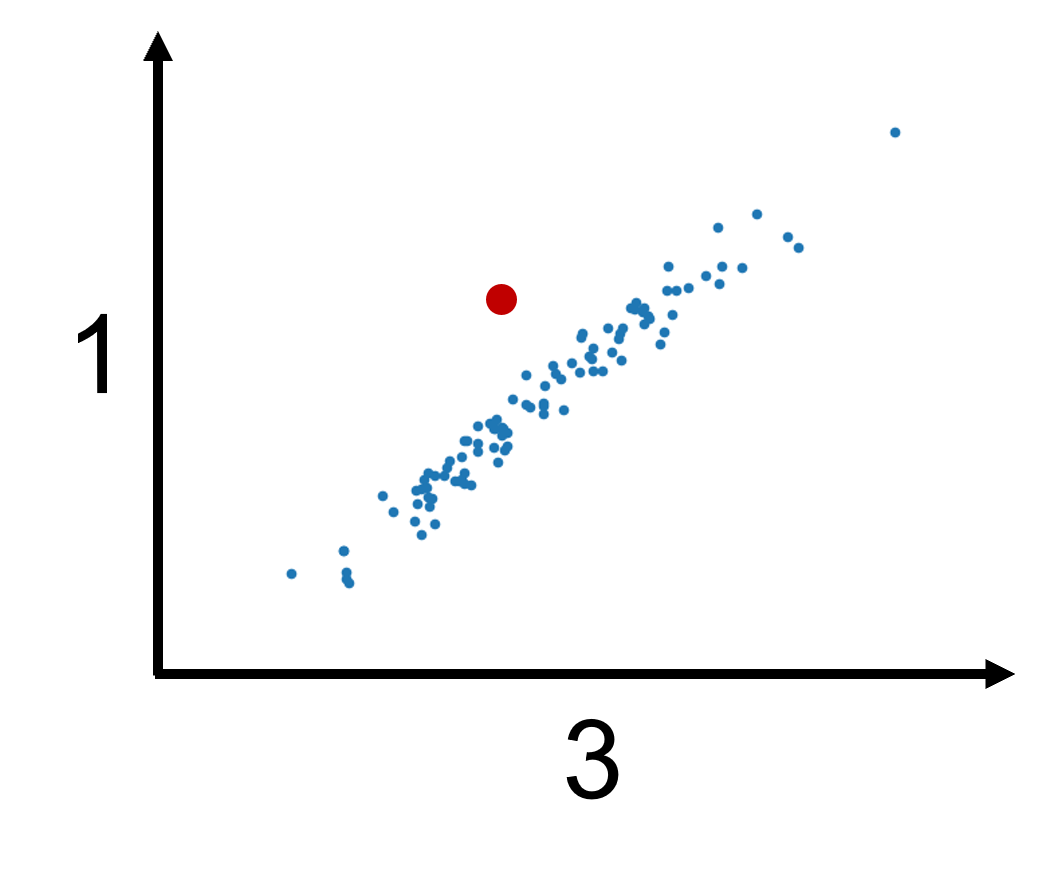}   
\includegraphics[width=0.25\textwidth]{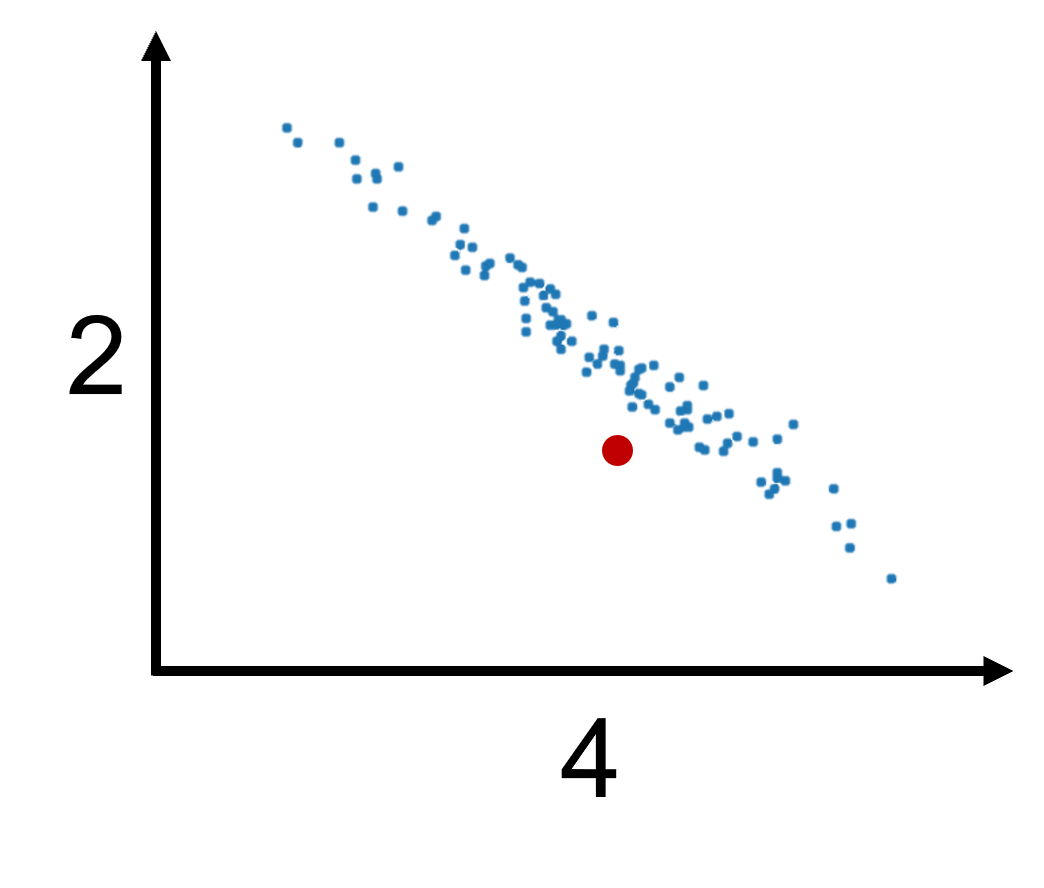} \\
(a) Initial order:
 $1$ \& $2$, $3$ \& $4$ &
(b) Correlated order:
 $1$ \& $3$, $2$ \& $4$  \\
\end{tabular}
\caption{
{\bf Reordering by correspondence.} 
Suppose we are given $100$ feature vectors, each with $4$ entries (features).
In the graphs, every axis represents one feature.
(a) shows that there is no correlation between features $1$ and $2$ (left), and similarly between features $3$ and $4$ (right).
Thus, the {\color{red}{anomalous red point}} cannot be distinguished from the  {\color{cyan}{normal cyan points}}.
However, after the features are reordered according to their correlation, it is much easier to distinguish between the anomalous point and the normal ones~(b).
}
\label{fig:reorder_importance}
\end{figure*}

\vspace{0.05in}
\noindent
{\bf Feature partition \& re-ordering.}
Evaluating the Eigen vectors by neighbors-of-neighbors (let alone locally) means that only a small number of points in the neighborhood of~$f$ is used for estimating the Eigen vectors.
This might be prohibitive since a feature vector of dimension $N$  cannot be estimated by $k << N$ neighbors, as it will result in major loss  of information.

To address this problem, we propose to estimate the Eigen vectors in parts.
We divide the vector features (entries) into equal-size sets and calculate the Eigen vectors for each set separately. 
In particular, we divide the feature vectors of dimension $N$ into at least $N/k$ sets.
In the following we denote the number of sets by $S$ and the dimension of the (sub)-feature vector of a set by $L$ (e.g. if $S=N/k$ then $L=k$).
In general, the more samples used to calculate the Eigen vectors, the larger $L$ may be.

Specifically, given a feature vector of the test point , $f$, and its $k$ nearest neighbors among the training samples, $f_i, 1\leq i \leq k$,  we partition $f$  and  $f_i$ into parts, $f_s$ \& $f_{i,s}$, $1<i<k, 1<s<S$.
We denote the Eigen vectors and  Eigen values associated with $f_i$, which are similarly partitioned, by $v_{ij,s}$, $e_{ij,s}$, $1\leq j\leq n$ $(n<L)$, respectively.

As before, we calculate the difference between  $f$ and each of its neighbors, however this time this is done per set.
The anomaly score of $f$, $AS$, takes into account the results of all the sets, as follows:
\begin{equation}
    AS(f) = \sum_{i=1}^{k}{\sum_{j=1}^{n}{\sum_{s=1}^{S}{|(f_{s} - f_{i,s})\cdot v_{ij,s}| \cdot \frac{1}{\sqrt{{e_{ij,s}}}}}}}.
    \label{eq:testi}
\end{equation}

The remaining question is how to partition the vectors into sets.
The disadvantage of using independent sets is that the relations between the features in the different sets are not taken into account.
This may be harmful when the anomalies depend on these relationships.
Figure~\ref{fig:reorder_importance} illustrates such a synthetic case, for vectors of $4$ features, where the red point is anomalous.
In a partition of the features into sets $\{1,2\}, \{3,4\}$, as in~(a), the red point is indistinguishable from the normal cyan points and thus will not be detected.

This problem can be mitigated by re-ordering the features before partitioning them into sets, based on the correlation between them.
If properly done, every set will contain the features that are most correlated to one another, resulting in more meaningful Eigen vectors. 
Figure~\ref{fig:reorder_importance}(b) illustrates the reordering effect, where the features are partitioned into sets $\{1,3\}, \{2,4\}$. 
When reordering is performed prior to splitting the features into sets, the red anomaly point is easily  spotted and distinguished from the normal cyan ones, and is thus detected as anomalous.

We propose to apply the following procedure for feature re-ordering.
First,  the correlations between all pairs of entries of all the feature vectors in the training set are computed. 
To maximize the correlation within each set, we re-order the feature vector entries of all the vectors simultaneously.
This is done in a greedy fashion as follows.
The first entry remains in place;
the second entry is switched with the one that is most correlated to the first.
From now on, until the number of features in the set is $L$, the subsequent entry is chosen as the one that has the highest average correlation with its previous two entries. 
When $L$ is reached, we start a new set, whose first entry  is chosen as the one that is least correlated with the last two features of the previous set.

\section{Experiments}
\label{sec:experiments}

We demonstrate the benefit of our method in two manners, 
In Section~\ref{subsec:improving} we replace the {\em k-NN} component of SoTA anomaly detection methods by our {\em k-NNN} operator and show improved results.
In Section~\ref{subsec:synthetic} we use our operator on structured synthetic features.
In both cases, it is demonstrated that even when applying our method on features extracted by networks that are not aimed at anomaly detection, the results gained are excellent.
For evaluation we use the the {\em AUROC} metric, which is the common evaluation of anomaly detection.

\subsection{Improving anomaly detection methods}
\label{subsec:improving}

In the following, we  replace the {\em k-NN}  component of SoTA {\em k-NN}-based anomaly detection methods by our {\em K-NNN} and evaluate the results on several datasets.

\vspace{0.05in}
\noindent
{\bf Networks \& datasets.}
We examine three 
systems that use {\em k-NN}:
(1)~k-NN applied to the features of ResNet18~\cite{targ2016resnet},
(2)~{\em Semantic pyramid anomaly detection  (SPADE)}~\cite{Cohen2020SubImageAD},
and (3)~{\em Panda}~\cite{reiss2021panda}.
We use four datasets:
(1)~{\em MVTec}~\cite{Bergmann2019CVPR}  contains $5,354$ high-resolution images of different object and texture classes.
It is divided into $3,629$ normal images for training and $1,725$ anomalous images for testing. 
The images contain more than $70$ different types of defects (anomalies), such as scratches, dents, and structural changes.
The ground truth was specified manually.
(2)~{\em Oxford flowers 102}~\cite{Nilsback08}  contains $102$ flower categories, with $1020$ training and validation images and $6,149$ test images, where each class includes $40$-$258$ images.
(3)~{\em Fashion MNIST}~\cite{xiao2017_online}  contains $10$ categories, with $60,000$ training and validation images and $10,000$ test images;  each class includes $6,000$  images.
(4)~{\em CIFAR10}~\cite{krizhevsky2009learning}   contains $10$ categories, with $50,000$ training and validation images and $10,000$ test images; each class includes $6,000$  images.

\begin{table}[tb]
    \centering
    \begin{tabular}{|l|c|c||c|c||c|c|}
    \hline
 Classes & Feature & Feature &  \cite{Cohen2020SubImageAD} &  \cite{Cohen2020SubImageAD} & \cite{reiss2021panda} & \cite{reiss2021panda}\\
  & +{\em k-NN} &  +{\em k-NNN} &   &  +{\em k-NNN} &  & +{\em k-NNN}\\
    \hline
\hline
carpet & 0.896 & \bf{0.990} & 0.928 & \bf{0.959} & 0.843 & \bf{0.898}\\
 \hline
grid & 0.444 & \bf{0.777} & 0.473 & \bf{0.663} & 0.554 & \bf{0.723}\\
 \hline

leather & 0.792 & \bf{0.986} & 0.954 & \bf{0.974} & 0.960 & \bf{0.975}\\
 \hline

tile & 0.986 & \bf{0.993} & 0.965 & \bf{0.970} & \bf{0.985} & 0.976\\
 \hline

wood & 0.636 & \bf{0.938} & 0.958 & \bf{0.985} & \bf{0.913} & 0.906\\
 \hline

bottle & 0.971 & \bf{0.983} & 0.972 & \bf{0.988} & \bf{0.992} & 0.982\\
 \hline

cable & 0.882 & \bf{0.934} & 0.848 & \bf{0.899} & 0.821 & \bf{0.863}\\
 \hline

capsule & 0.803 & \bf{0.919} & 0.897 & \bf{0.941} & 0.911 & \bf{0.919}\\
 \hline

hazelnut & 0.903 & \bf{0.991} & 0.881 & \bf{0.966} & 0.925 & \bf{0.968}\\
 \hline

metal\_nut & 0.813 & \bf{0.913} & 0.710 & \bf{0.857} & 0.788 & \bf{0.860}\\
 \hline

pill & 0.738 & \bf{0.882} & 0.801 & \bf{0.822} & 0.757 & \bf{0.786}\\
 \hline

screw & 0.712 & \bf{0.840} & 0.667 & \bf{0.839} & 0.690 & \bf{0.805}\\
 \hline

toothbrush & 0.886 & \bf{0.969} & 0.889 & \bf{0.953} & 0.861 & \bf{0.914}\\
 \hline

transistor & 0.878 & \bf{0.936} & 0.903 & \bf{0.929} & 0.871 & \bf{0.902}\\
 \hline

zipper & 0.937 & \bf{0.964} & \bf{0.966} & 0.949 & 0.934 & \bf{0.951}\\
 \hline

mean & 0.819 & \bf{0.934} & 0.854 & \bf{0.913} & 0.854 & \bf{0.895}\\
 \hline
    \end{tabular}
    \caption{
   {\bf  Replacing the {\em k-NN} component by our {\em k-NNN} on MVTec.}
Our {\em k-NNN} improves the mean performance of all networks, as well as the performance for almost all the classes.
}
\label{tbl:quantitative_mvtec}
\end{table}

\begin{table}[tb]
    \centering
    \begin{tabular}{|c|c|c||c|c||c|c|}
    \hline
 Dataset & Feature & Feature &  \cite{Cohen2020SubImageAD} &  \cite{Cohen2020SubImageAD} & \cite{reiss2021panda} & \cite{reiss2021panda}\\
  & +{\em k-NN} &  +{\em k-NNN} &   &  +{\em k-NNN} &  & +{\em k-NNN}\\
    \hline
\hline
CIFAR10 & 0.841 & \bf{0.871} & 0.893 & \bf{0.922} & 0.939 & \bf{0.943}\\
 \hline

Fashion  & 0.935 & \bf{0.936} & 0.911 & \bf{0.919} & 0.954 & \bf{0.958}\\
 \hline

Flowers & 0.615 & \bf{0.895} & 0.917 & \bf{0.919} & 0.935 & \bf{0.944}\\
 \hline
    \end{tabular}
    
    \caption{
   {\bf  Replacing  {\em k-NN}  by our {\em k-NNN} on three additional datasets.} 
Our AUROC results outperform those of the $3$ methods.
}
\label{tbl:quantitative_multi_datasets}
\end{table}

\vspace{0.05in}
\noindent
{\bf Results.}
Table~\ref{tbl:quantitative_mvtec} shows   the results on MVTec.
In this dataset, every class has anomalous examples of its own.
Our method improves the mean performance of all three networks.
Furthermore, it improves the performance for almost all the classes (except $3$ classes for a single network).

Table~\ref{tbl:quantitative_multi_datasets} reinforces the above results.
It shows improved performance   on  three additional datasets, which differ greatly from one another, in their size, type and diversity.

Table~\ref{tbl:diverse-normal} further tests our model on highly diverse normal classes.
In particular, in this experiment we define as normal the images from all the classes of a specific dataset, except for one (note that no classification is needed beforehand).
Thus, only images from a single class should be detected as anomalous.
The table shows that though the performance of all networks suffers when the normal set is diverse, our {\em k-NNN} still improves the results.
In fact, it usually improves the results more.

Table~\ref{tbl:increasing} further studies the issue of diversity.
In MVTec, every class has its own normal and anomalous examples, hence it may be considered as a set of independent datasets.
In this experiment, we gradually increase the number of classes, i.e. if the number of classes is~$5$, we consider all the normal (unclassified) images of the~$5$  classes as normal and all the anomalous (unclassified) images of these classes as anomalous.
(Table~\ref{tbl:quantitative_mvtec} is the base case.)
The table  shows that generally the more diverse the normal class is, the more advantageous our method is.
This is not surprising, as after all this is exactly what  {\em k-NNN} is supposed to do---be adaptive to the structure of the feature space.

\begin{table}[tb]
    \centering
    \begin{tabular}{|c|c|c||c|c||c|c|}
    \hline
 Dataset & Feature & Feature &  \cite{Cohen2020SubImageAD} &  \cite{Cohen2020SubImageAD} & \cite{reiss2021panda} & \cite{reiss2021panda}\\
  & +{\em k-NN} &  +{\em k-NNN} &   &  +{\em k-NNN} &  & +{\em k-NNN}\\
\hline
\hline
CIFAR10 & 0.662 & \bf{0.719} & \bf{0.694} & 0.667 & 0.599 & \bf{0.610}\\
\hline

Fashion  & 0.760 & \bf{0.780} & 0.729 & \bf{0.739} & 0.683 & \bf{0.698}\\
 \hline

Flowers & 0.612 & \bf{0.681} & 0.624 & \bf{0.686} & 0.668 & \bf{0.689}\\
 \hline

 \hline
\end{tabular}
\caption{
   {\bf  Performance on diverse normal sets.}
   Images from all the categories, expect one, are considered  normal, and images from that single class are anomalies.
   The AUROC  results are averaged across all the classes, i.e. each class is considered anomalous once.
   When replacing  {\em k-NN}  by our {\em k-NNN} in various networks, our operator is usually more beneficial than for homogeneous sets. 
}
\label{tbl:diverse-normal}
\end{table}

\begin{table}[tb]
    \centering
    \begin{tabular}{|c||c|c||c|c||c|c|}
    \hline
 \#normal & Feature & Feature &  \cite{Cohen2020SubImageAD} &  \cite{Cohen2020SubImageAD} & \cite{reiss2021panda} & \cite{reiss2021panda}\\
 classes & +{\em k-NN} &  +{\em k-NNN} &   &  +{\em k-NNN} &  & +{\em k-NNN}\\
\hline
\hline
 \hline
5 & 0.806 & \bf{0.890} & 0.760 & \bf{0.938} & 0.599 & \bf{0.783}\\
 \hline
7 & 0.743 & \bf{0.852} & 0.716 & \bf{0.913} & 0.597 & \bf{0.817}\\
 \hline
11 & 0.680 & \bf{0.760} & 0.747 & \bf{0.901} & 0.631 & \bf{0.809}\\
 \hline
15 & 0.627 & \bf{0.757} & 0.691 & \bf{0.884} & 0.627 & \bf{0.814}\\

 \hline
\end{tabular}
\caption{
   {\bf  Performance when increasing the normal sets.}
    Out of the $15$ highly diverse classes of MVTec, 
    we use an increasing number of sets, of which all their normal and anomalous images are considered as such.
   As before, no classification is performed beforehand.
Our operator is especially beneficial on divese sets.
}
\label{tbl:increasing}
\end{table}

\subsection{Performance on synthetic benchmarks}
\label{subsec:synthetic}

\begin{figure}[t]
\centering
\begin{tabular}{ccc}
\includegraphics[width=0.17\textwidth]{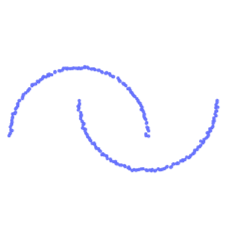} &
\includegraphics[width=0.15\textwidth]{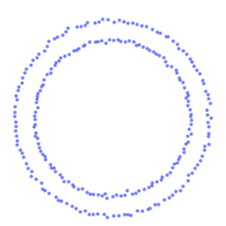} &
\includegraphics[width=0.16\textwidth]{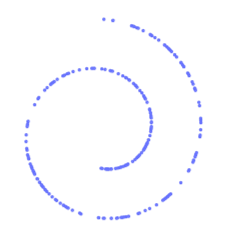} \\
(a) Moons & (b)  Circles &
(c)  Swiss roll 
\end{tabular}
\caption{
{\bf  Synthetic benchmarks}
}
\label{fig:synthetic}
\end{figure}

We study the performance of our method on synthetic well-thought benchmarks, which are frequently used for visualizing clustering and classification algorithms ~\cite{scikitlearn}.
These datasets demonstrate the strength of our method when the embedding space is structured.
These benchmarks are created by various random sampling generators, which enable to control their size and their complexity.
Figure~\ref{fig:synthetic} illustrates the three benchmarks we use: 
\begin{enumerate}
    \item 
    {\bf Moons.} The points are  arranged in two interleaving half circles.
    \item
    {\bf Circles.} The points are arranged in two circles, with the same center point but different radii.
    \item
    {\bf  Swiss roll.}
    The points are arranged in a rolled-up shape, similar to a Swiss roll pastry.
\end{enumerate}

In our setup, we consider all the generated points as embeddings of normal examples.
The farthest a point in the plane is from the specific distribution, the more anomalous it should be.
For the training, half of the generated points were used.
For the evaluation, the other half was considered as the true positives.
We generated the negatives (anomalies) by uniformly sampling the plane.
In our experiments we used $100$-$500$ points for training and $5,000$ for  testing.

 Table~\ref{tbl:synthetic-results} shows the benefit of our operator quantitatively.
 Note that we cannot compare against other SoTA methods, as they compute the embedding as an integral part of the network, whereas here the embedding is given. 
Figure~\ref{fig:synthetic-results} illustrates the results qualitatively, showing that the classical {\em k-NN} erroneously captures a wide area around the curve as normal; the global {\em k-NN} identifies in-curve points as anomalous (e.g., the spaces in the spirals); the local {\em k-NN} adds anomalous curves, as seen in the case of the moons and the roll.
In contrast, our method accurately captures the thin  normal curves, including the holes in them and their continuation.
We define these methods below.

 \begin{table}[tb]
    \centering
    \begin{tabular}{|l|c|c|c|c|}
    \hline
     Dataset &  k-NN &  Local &  Global & {\em k-NNN} \\
  \hline
\hline
Moons & 0.8214 & 0.8408 & 0.8860 & \bf{0.9162}\\
 \hline
Circles & 0.8079 & 0.8123 & 0.8109 & \bf{0.8163}\\
 \hline
Swiss roll & 0.9496 & 0.9496 & 0.9794 & \bf{0.9816}\\
 \hline
    \end{tabular}
    \caption{
    {\bf Performance on the synthetic dataset.}
    This table shows that our {\em k-NNN} method outperforms all variants of  {\em k-NN}.
    }
\label{tbl:synthetic-results}
\end{table}

\section{Ablation study}
\label{subsec:ablation}

\subsection{{\em k-NN}  methods }
In this section we study variants of  {\em k-NN} normalization methods and compare them to  our  {\em k-NNN} operator. 

\begin{figure}[tb]
  \centering
     \begin{tabular}{cccc}   
    \includegraphics[height=0.1\textwidth]{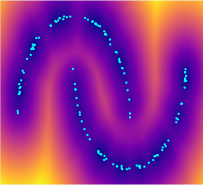}&
    \includegraphics[height=0.1\textwidth]{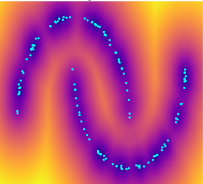}&
    \includegraphics[height=0.1\textwidth]{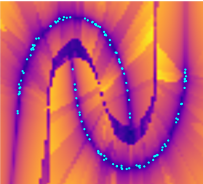}&
    \includegraphics[height=0.1\textwidth]{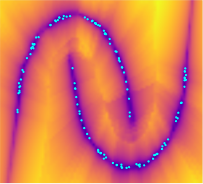} \\
   \includegraphics[height=0.1\textwidth]{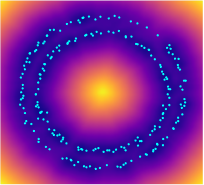}&
    \includegraphics[height=0.1\textwidth]{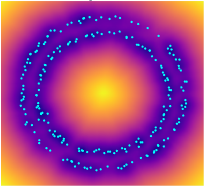}&
    \includegraphics[height=0.1\textwidth]{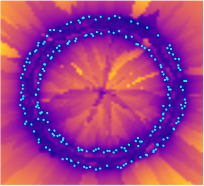}&
    \includegraphics[height=0.1\textwidth]{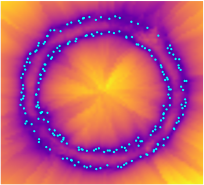} \\
     \includegraphics[height=0.1\textwidth]{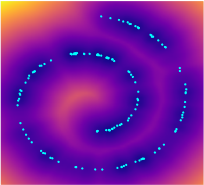}&    \includegraphics[height=0.1\textwidth]{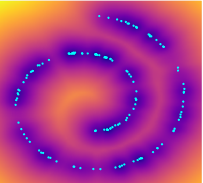}&    \includegraphics[height=0.1\textwidth]{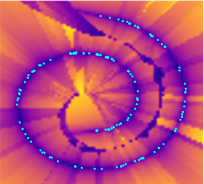}&
    \includegraphics[height=0.1\textwidth]{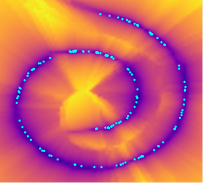} \\
         (a) K-NN & (b) Global & (c) Local &  (d) K-NNN 
    \end{tabular}
    \caption{
    {\bf Qualitative results on synthetic datasets.}
    Only our method captures the curve distributions accurately.
    That is to say, the only blue regions across all datasets are the thin curves, including the holes in them and their continuation.
  }
  \label{fig:synthetic-results}
\end{figure}

\vspace{0.05in}
\noindent
{\bf 1. Baseline (no normalization).}
The Euclidean distance between the feature vectors is used without normalizing the features, as done in~\cite{breunig2000lof,Cohen2020SubImageAD,knorr2000distance,ramaswamy2000efficient}. 
The advantage of no normalization is speed and having a single parameter ($k$, the number of neighbors).
However, ignoring the shape of the local and global embedding space might harm performance.
This is illustrated in Figures~\ref{fig:dots_normalization},\ref{fig:synthetic-results}(a), where the normal points within the holes and in the continuation of the curve are erroneously detected as anomalous (the red rectangles in Figures~\ref{fig:dots_normalization}(a)).

\vspace{0.05in}
\noindent
{\bf 2. Global Eigen-vector normalization.}
 All the training points are used to calculate the Eigen vectors $\{v_1, v_2,\dots v_n\}$ and their associated Eigen values $\{e_1, e_2,\dots e_n\}$.
At test time, given an image represented by its feature vector~$f$, we compute its $k$ nearest neighbors $\{f_{1}, f_{2},\dots f_{k}\}$ from the training set and  normalize 
$(f-f_{i})$, based on the Eigen values.
The anomaly score,  $AS$, is:
\begin{equation}
    AS(f) = \sum_{i=1}^{k}{\sum_{j=1}^{n}{|(f-f_{i})\cdot v_j| \cdot \frac{1}{\sqrt{{e_j}}}} }.
    \label{eq:scoref}
\end{equation}
As before, normalization is performed by the square  root of the inverse of the Eigen value,  in accordance with our observation that lower Eigen values are indicative of an anomaly.
Note that unlike Eq.~\ref{eq:testf}, here all the vectors in the embedding space are normalized in the same manner.

The advantages of global normalization are that it is less affected by noise than considering only a subset of points and that the relations between the features are taken into account.
Figures~\ref{fig:dots_normalization},\ref{fig:synthetic-results}(b) show that global normalization improves upon the baseline {\em k-NN}.
However, global normalization ignores the different directions for different regions of the embedding space.
As illustrated in the red square of Figure\ref{fig:dots_normalization}(b),  the continuation of the curve is considered anomalous, though it should be normal.
This is due to the fact that the global Eigen vector direction does not match the correct local direction.

\vspace{0.05in}
\noindent
{\bf 3. Local normalization.}
In local normalization, we calculate the Eigen vectors for each test point  in the embedding space locally, based on its $k$ nearest neighbors in the train set. 
Then, as before, we calculate the difference between the test sample $f$ and each of its neighbors, and apply Eq.~\ref{eq:scoref}.

Figures~\ref{fig:dots_normalization},\ref{fig:synthetic-results}(c) show that indeed all the points along the curve are correctly detected as normal
(e.g, see the green squares in Figure~\ref{fig:dots_normalization}(c)).
However, artifacts are created---the blue "snakes" in-between the arcs of the moons (partially visible in the red rectangle), within the roll and between the circles.
The points on these snakes should be classified as anomalous, whereas they are considered to be normal.

\vspace{0.05in}
\noindent
{\bf Evaluation with different embeddings.}
In Table~\ref{tbl:anomaly_models_comp} we further compare these approaches using a variety of image embeddings.
Specifically, we used 
ResNet~\cite{targ2016resnet}, ResNeXt~\cite{xie2017aggregated}, and  Dino-ViT-S/8~\cite{caron2021emerging}).
For each embedding, we applied the nearest neighbor variants to detect anomalies on MVTec.
It shows that our {\em k-NNN} outperforms all other variants.
These results are consistent with those presented in Table~\ref{tbl:synthetic-results}, where the same experiment was
performed on the 2D synthetic dataset of~\cite{Scikit-learn}.
It is interesting to note that this very simple method
already manages to detect anomalous images pretty well.

\begin{table}[tb]
    \centering
    \begin{tabular}{|l|c|c|c|c|}
    \hline
    Network &  k-NN &  Local &  Global & {\em k-NNN} \\
   \hline
\hline
    \multicolumn{5}{|c|}{Max}  \\
 \hline
Dino-Vits8~\cite{caron2021emerging} & 0.9357 & 0.9357 & 0.9510 & \bf{0.9556}\\
 \hline
Resnet50~\cite{targ2016resnet} & 0.7690 & 0.7693 & 0.8040 & \bf{0.8095}\\
 \hline
Resnet101~\cite{targ2016resnet} & 0.7690 & 0.7693 & 0.8040 & \bf{0.8095}\\
 \hline
ResNext50~\cite{xie2017aggregated} & 0.7690 & 0.7693 & 0.8040 & \bf{0.8095}\\
 \hline
ResNext101~\cite{xie2017aggregated} & 0.7690 & 0.7693 & 0.8040 & \bf{0.8095}\\
 \hline
\multicolumn{5}{|c|}{Mean}  \\
  \hline
Dino-Vits8~\cite{caron2021emerging} & 0.9194 & 0.9207 & 0.9358 & \bf{0.9379}\\
 \hline
Resnet50~\cite{targ2016resnet} & 0.7282 & 0.7283 & 0.7205 & \bf{0.7350}\\
 \hline
Resnet101~\cite{targ2016resnet} & 0.7256 & 0.7257 & 0.7414 & \bf{0.7423}\\
 \hline
ResNext50~\cite{xie2017aggregated} & 0.7282 & 0.7283 & 0.7222 & \bf{0.7382}\\
 \hline
ResNext101~\cite{xie2017aggregated} & 0.7284 & 0.7285 & 0.7374 & \bf{0.7427}\\
 \hline
    \end{tabular}
    \caption{
    {\bf Comparison of various {\em k-NN}  methods.}
    Our {\em k-NNN} operator outperforms all other nearest neighbor variants on MVTec.
    Furthermore, simply finding the embedding using Dino-Vits8 and then running our operator detects anomalous images pretty well.
    }
\label{tbl:anomaly_models_comp}
\end{table}

\subsection{Parameters and runtime}
\noindent
{\bf How many neighbors should be used?}
For clarity, throughout the paper, we did not elaborate on having two neighboring parameters: the number of neighbors of a given test image and the number of neighbors of the train images, which are pre-computed and stored (i.e., neighbors of neighbors).
Table~\ref{tbl:quantitative_multi_NoN} shows typical results (here, $15$ classes from Table~\ref{tbl:increasing}).
It shows that
it  is beneficial to use a small number of neighbors and a large number of neighbors of neighbors.
For instance, having $3$ neighbors, each having $25$ neighbors, is superior to having $75$ direct neighbors (the case of {\em k-NN}), improving the performance from $0.616$ to $0.757$.
Intuitively, a few nearby neighbors and enough of their neighbors suffice to provide good statistics of the nearby space.
This justifies the key idea of the paper: {\em k-NN}, which considers only Euclidean distances, cannot capture the structure of the space, even if more neighbors are added.
Conversely, our {\em k-NNN} captures the space structure and addresses the problem of an anomaly being closer to certain clusters than normal examples from each other.
In our implementation we use  $3$ neighbors and  $25$ neighbors of neighbors across all datasets.

\begin{table}[tb]
    \centering
\begin{tabular}{|c|c|c|}
\hline
 \#neighbors & \#neighbors-neighbors & performance \\
\hline
\hline
  1 &   75 &  0.753 \\
  3 &   20 &  0.753 \\
  3 &   25 &  {\bf 0.757} \\
  3 &   75 &  0.755 \\
  4 &   20 &  0.754 \\
  5 &   15 &  0.753 \\
 10 &    5 &  0.686 \\
 \hline
 60 &    0 &  0.616 \\
 75 &    0 &  0.616 \\
 80 &    0 &  0.634 \\

    \hline
    \end{tabular}
    \caption{{\bf How many neighbors are needed?}
    Considering a few nearby neighbors and many of their neighbors (top) is advantageous to having many neighbors (bottom).
    This verifies the key idea of the paper---neighbors-of-neighbors (top) capture the structure of space  much better than only neighbors do (bottom).
}
\label{tbl:quantitative_multi_NoN}
\end{table}

\vspace{0.05in}
\noindent
{\bf Sub-feature vector dimension.}
Another parameter that should be set is $L$, the dimension of the sub-feature vector used for the partition, which
might affect the algorithm's performance and runtime.
We used  $L=5$, which experimentally exhibited the best performance.
For instance, in Table~\ref{tbl:increasing}~\cite{Cohen2020SubImageAD}, when $L=4$ the performance already decreased by $0.006$ and similarly when using larger $L$.

\vspace{0.05in}
\noindent
{\bf Runtime.}
A fundamental advantage of our method is that no training is needed. 
We use the features generated by any network and apply our {\em k-NNN} operator.
If the Eigen vectors
are computed during pre-processing, the  inference runtime is instantaneous. 
In particular,
computing the Eigen vectors and the partition  during pre-processing takes about $0.074$ seconds per image.
Given a test image, it takes $0.014$ second to determine  anomaly, when using $3$ neighbors and $25$ neighbors of neighbors.
The experiments are performed  on the CPU ({\em AMD EPYC 7763}).

\vspace{0.05in}
\noindent
{\bf Limitations.}
The disadvantage of our {\em  k-NNN} is its running time during pre-processing and the memory needed to store the Eigen vectors.
Furthermore, our operator has   $3$  hyper-parameters that need to be to tuned, in comparison to a single parameter in classical {\em k-NN}.

\section{Conclusion}
\label{sec:conclusion}

This paper has proposed a new nearest-neighbor  operator, \em{k-NNN}\em{}, which leverages neighbors of neighbors statistics.
The underlying idea in that these statistics provide information regarding the shape of the feature space.
Our operator computes and stores the Eigen vectors of the train set neighbors.
During inference, these vectors are used to compute a more accurate anomaly score, utilizing a novel normalization scheme.
Additionally, we suggest to compute these Eigen vectors in parts, using multiple features sets. 
This addresses the problem of how to estimate a vector in high dimension with insufficient number of neighbors.

We showed that multiple anomaly detection networks can be improved by simply replacing their {\em k-NN component} by our {\em K-NNN}, both in homogenous datasets and in diverse datasets.

\vspace{0.1in}
\noindent
{\bf Acknowledgement.}
This work was supported by the
Israel Science Foundation 2329/22.

\newpage
\bibliographystyle{splncs04}
\small
\bibliography{citation.bib}

\end{document}